# An Approach for Thyroid Nodule Analysis Using Thermographic Images

**J.R. González, É.O. Rodrigues, C.P. Damião, C.A.P. Fontes, A.C. Silva, A.C. Paiva, H. Li, C. Du and A. Conci**

**Abstract** Thyroid cancer is said to be the second most common type of cancer in female individuals and the third in males by 2030, according to projections. In general, detecting cancer in its early stages improves the chance of survival of the individual. Thermography is a diagnostic tool that has been increasingly used to detect cancer and abnormalities, including that of thyroid. Various methods to segment and detect hot regions in thermograms and, consequently, to detect suspicious tissues present in these images have been proposed. It is well known that medical diagnosis yields a great deal of information. Thus, physicians have to comprehensively analyse and evaluate this information in a short period of time, which is infeasible in most cases. In this work, we perform a general review of thermography, focusing on the thyroid analysis. We propose protocols for image acquisiton and an autonomous registration for thyroid images. We also perform analyses of the image data, which include feature extraction, image processing, and

J.R. González · É.O. Rodrigues (✉) · A. Conci
Computer Science Department, Universidade Federal Fluminense,
Av. Milton Tavares de Souzas, s/n—Boa Viagem, Niterói, RJ 24210-330, Brazil
e-mail: erickr@id.uff.br

J.R. González
e-mail: jgonzalez@ic.uff.br

A. Conci
e-mail: aconci@ic.uff.br

C.P. Damião · C.A.P. Fontes
Radiology Department, Hospital Universitário Antônio Pedro (HUAP),
Av. Marquês do Paraná, 303, Niterói, RJ, Brazil
e-mail: charbeldamiao@yahoo.com.br

C.A.P. Fontes
e-mail: cfontes@id.uff.br

A.C. Silva · A.C. Paiva
Applied Computation Group NCA-UFMA, Universidade Federal do Maranhão,
Av. dos Portugueses, São Luís, 1966, MA, Brazil
e-mail: ari@dee.ufma.br

a possible approach for classification of healthy or unhealthy patients. In summary, this work presents a pilot project for detection of tumors in our university hospital, which is part of an effort to support preventive medical actions in our endocrinology department. Under some future adjustments, this project will be submitted for approval by the ethics and research committee of Hospital Universitário Antonio Pedro at Universidade Federal Fluminense (HUAP-UFF) and to the Brazilian Ministry of Health Ethical committee under the name: Evaluation of the importance of thermography to aid diagnosis of thyroid nodules of patients in HUAP-UFF (in Portuguese: Avaliação da importância da termografia no auxílio à investigação diagnóstica de nódulos tireoidianos em pacientes acompanhados no HUAP-UFF).

**Keywords** Thyroid cancer · Neck nodule · Thermography · Image registration · Classification · Threshold · Sobel · Image analysis

## 1 Introduction

Equipment used in medical investigation usually rely on local changes in body parameters like density, sound propagation, electromagnetic resonance, etc. Healthy and diseased areas of the human body can be analysed and compared when using these equipment. In general, features extracted from both healthy and diseased areas are usually processed and ultimately compared to infer a diagnosis. These features can be anything that stems from the information encapsulated by these images. However, when primary features such as the physical information are compared directly, it yields low recognition rates [1]. In contrast, more robust features are usually more valuable, e.g., features that extract information from an area of the image instead of a single point, since areas are analysed as a whole.

The temperature distribution on the human skin can be captured with a proper camera, resulting in a thermal image (Fig. 1). These images represent temperature patterns of the body, which are highly symmetric with respect to the vertical axis of

A.C. Paiva
e-mail: paiva@deinf.ufma.br

H. Li
School of Mathematical Sciences, Shandong Normal University,
88 East Wenhua Road, Jinan, Shandong 250014, People's Republic of China
e-mail: lihl@sdnu.edu.cn

C. Du
School of Mathematical Sciences, University of Jinan, 336 West Road
of Nan Xinzhuang, Jinan, Shandong 250022, People's Republic of China
e-mail: sms_duchb@ujn.edu.cn

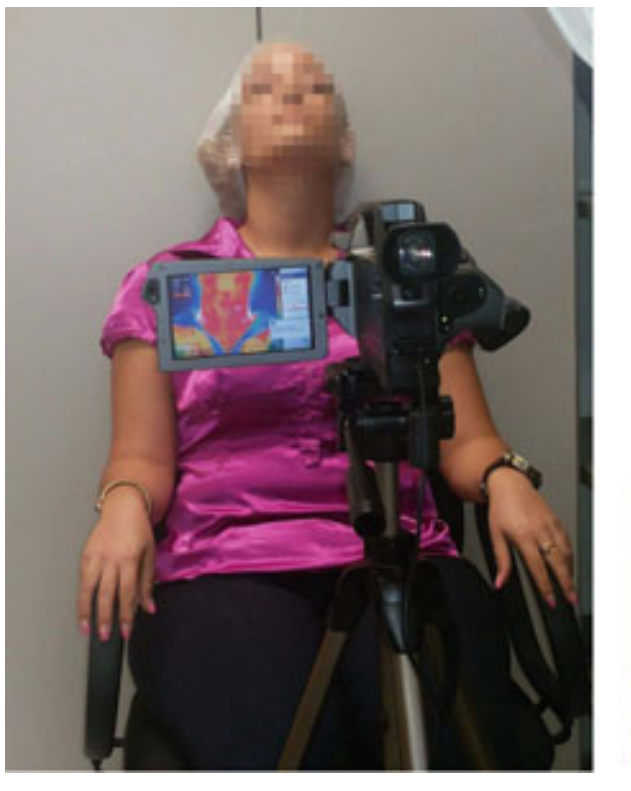

(a) Thermography acquisition.

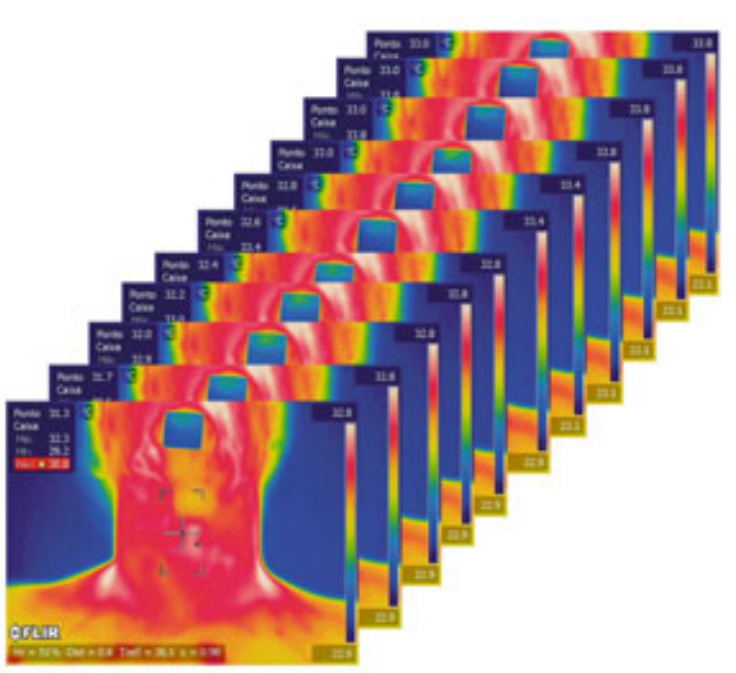

(b) Infrared image sequencing.

**Fig. 1** Thermographic imaging

the sagittal plane. Variations of this infrared map on serial imaging (images taken over time) can also evidence a sign of abnormality. Thermography detects temperature modifications in tissues that appear before or during many diseases, including cancer. Besides, it also detects physiologic or functional changes, but it does not have the ability to pinpoint the internal location of the problem. However, the Pennes equation [2] can be used to describe the contributions of the internal body elements to the skin thermal distribution. In such equation, the heat transfer problem is written in a simplified form. However, the thermal conductivity of tissues, their density, the specific heat of the blood, blood perfusion rate, metabolic heat generation rate, core temperature of the body, arterial blood temperature, environment and local temperature of each tissue must be known a priori.

Artificial intelligent methods can also be used to predict and find abnormalities in images. In this case, a reasonable number of exams with proven diagnosis, i.e., ground truth, is usually required. These labelled exams are used by a classification algorithm as source of information for finding patterns. The found patterns are used to classify patients whose diagnoses are unknown. In order to be consistent in this process, the same acquisition protocol, conditions and computer processing techniques should be respected for every patient.

The methodology to capture the images can either consider a (1) dynamic acquisition protocol [1], where several images are acquired as a time series, or a (2) static protocol, where one single image is acquired [3]. In the dynamic protocol, the area to be examined is cooled by an air stream before the examination, which changes the patient body temperature. Meanwhile, a series of images are acquired while their body return to the thermal equilibrium with the environment (Fig. 1b shows a time series).

During the acquisition process, the patients breathe (inhale and exhale) performing some small movements. These movements change the coordinates of points of their body in relation to a previously acquired image. Figure 2 shows the exact movement of keypoints over the time series acquisition. Figure 2a shows a common point located at the patient chin, and Fig. 2b shows a common point located at the patient's neck. Errors can be propagated to the subsequent computational analyses if these errors are not corrected. Therefore, approaches to relate these points should be one of the first steps in the computational analysis of these thermograms.

It is possible to correct these differences using a process named image registration [4], which is more extensively addressed in the next section. Registrations are based on global or local transformations and are fundamental in time series acquisition [5]. Image registration usually matches the maximum number of common points in the images using various types of transformations respecting rigid bodies (translation and rotation) or nonrigid bodies (transformations that do not preserve the distances between two points) [6].

The use of very generic or incorrect transformations increases the overall processing time and may not even produce adequate registered images. To achieve an efficient registration, the movements that the patients perform must be carefully analysed. A function that is sufficient to correct the performed movements must be known a priori to improve the processing time. For instance, there is no need to apply a complex warping or perspective transformation if a simple translation is enough, which is also far less complex.

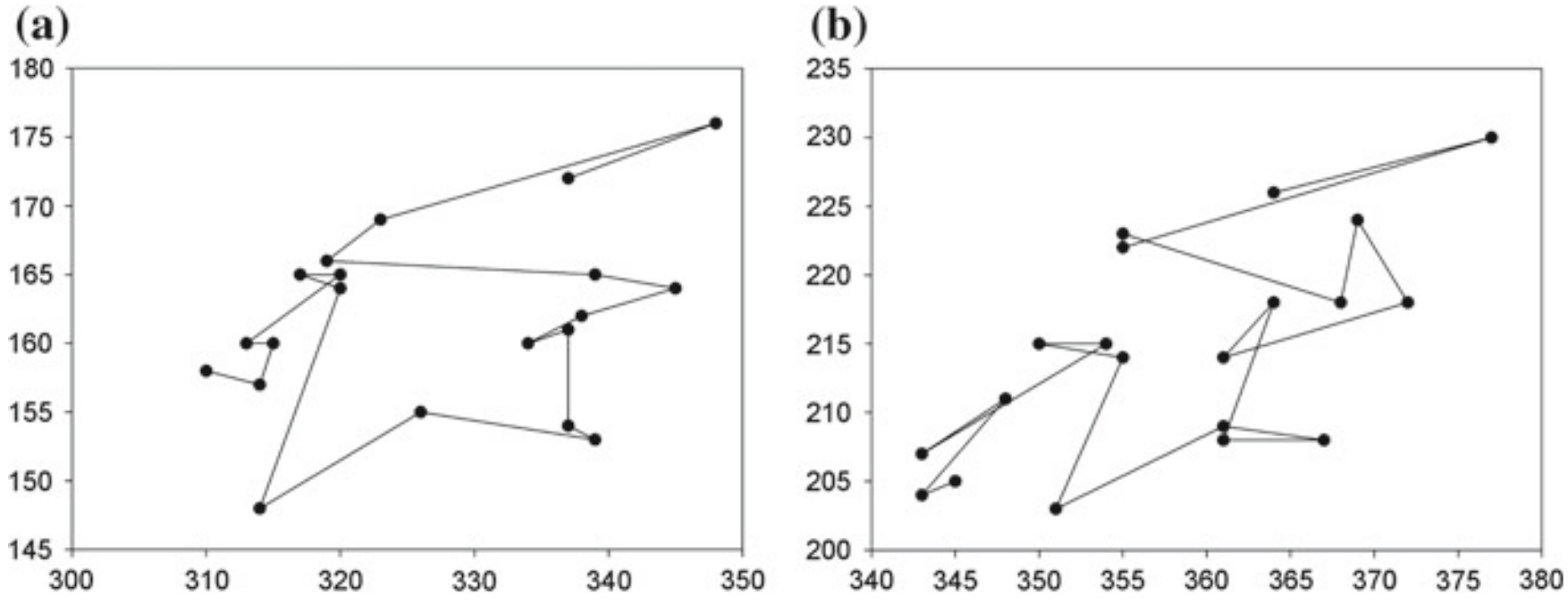

**Fig. 2** Displacement of common keypoints over the time series acquisition

## 2 Literature Review

Thermographic cameras are built with sensors that capture infrared (IR) rays. The camera sensor is able to perceive radiation at temperatures above absolute zero. All objects that have temperatures above this value (i.e., 0 K or −273 °C) emit infrared radiation. The measured infrared radiation emitted by one point of the skin can be converted directly to a temperature value that represents this point.

Some diseases trigger physiological and biological processes that interfere with the human temperature distribution. Nitric oxide, for instance, which is produced by cancer cells, interferes with the normal neuronal control of blood vessel flows, causing a local vasodilation in the early stages of cancerous growth, and also enhancing the angiogenesis in later stages. Increased blood flows can raise the temperature of the related area. Deep lesions seem also to have the ability to induce changes in the skin temperature. Cancerous metabolic processes also seem to contribute to the detectable heat increase.

Among other approaches, IR thermography can detect a disease or abnormality by identifying temperature differences on the symmetric areas being compared. Physiological changes and the development of neoangiogenesis are often associated to these asymmetric temperatures distributions. Asymmetry between symmetric areas is one of the most important indicators that can be quantified. Locating areas within the region of interest (ROI) that contain a high degree of blood perfusion or vessels is also of great importance. Thermogram interpretation is based, in most occasions, on exaggerate vascularisation, hot spots and on asymmetries between the said symmetric areas.

The next sections of this work present a small survey on infrared imaging for thyroidal screening and addresses related works in the light of their applications in computer vision and medicine. Consequently, the ideas in the following two sections are organized according to the standard flowchart of pattern recognition systems. The entire process is composed by: (1) image acquisition protocols, (2) storage, (3) Region of Interest (ROI) and registration, (4) segmentation methods, (5) feature extraction, (6) classification or diagnosis.

### *2.1 Acquisition Protocol*

A protocol for thermal imaging can be categorized as static or dynamic, based on the behaviour of the body in relation to heat transfer. In the static acquisition, the body of the patient is thermally stabilized with the environment. On the other hand, static acquisition relies on the thermal equilibrium of the patient with the environment. Dynamic acquisition is applied to monitor the recovery of the skin temperature after an induced thermal stress (e.g., heating or cooling) or chemical stress (e.g., vasoconstrictions or vasodilations).

The type of acquisition can also be categorized according to the repetition as single, sequential or accompanied capture. In single acquisitions, an image of the patient is captured in an instant of time. This type of acquisition is suitable to identify cold and hot spots and for measuring the asymmetries on skin temperature. In the sequential mode, however, a series of images are acquired sequentially. Finally, in the accompanied mode, the acquired images are separated by a substantial time interval (usually 3 or 6 months) in order to follow the progress of some disease and also to detect it early (i.e., to detect changes in the patterns early).

The influence of the room temperature, as well as the stabilization time of the body in relation to this temperature was investigated by Usuki et al. [7]. In dynamic acquisitions with cold stress protocols, the expected pattern is a balance between heat conduction from tissues and deeper vessels, and heat loss by radiation and air convection at the surface. Beyond the image acquisition, some important aspects should be regarded as well, such as date of examination and patient's age, as well as use of drugs or hormonal therapy.

An alternative acquisition protocol was evaluated by Agostini et al. [8]. The proposed protocol aims to observe the influence of blood perfusion and skin temperature fluctuation. The authors acquire a sequence of consecutive thermal images with rate ranging from 50 to 200 frames/s. Thereafter, they use the frequency domain of the small temperature fluctuation in the studied area, rather than considering the classical static skin temperature. Kapoor et al. [9] recommend their patients to stop smoking for 2 hours before the test, to avoid alcohol and caffeine, and to not apply lotions on the acquisition area. Ng et al. [3] do also orient patients to abstain from any physical activities for 20 min before the exam, in order to reduce the body metabolism and stabilize the body temperature.

## 2.2 Registration

Image registration is a mapping of points that relate two images where one is usually called reference or fixed image and the other is usually called sensible or moving image. Registration is an optimization problem where the goal is to maximize the correspondence of points from the reference image to the ones in the sensible image using transformations [10]. Figure 3 shows an overall example of a registration.

A desirable registration or transformation $T$ is achieved when the statement $T$ *(SensibleImage)* $\cong$ *ReferenceImage* is true. The symbol $\cong$ represents an approximation and is subjective. The comparison depends on the similarity measure chosen to compare the reference and transformed images. After the transformation, the image is converted from the function domain to a raster. That is, rasterization is the defined as the process of discretizing and painting pixels from the function domain to a grid (e.g., an image, screen, etc). Interpolations are required for doing so. The entire registration process is shown in Fig. 4.

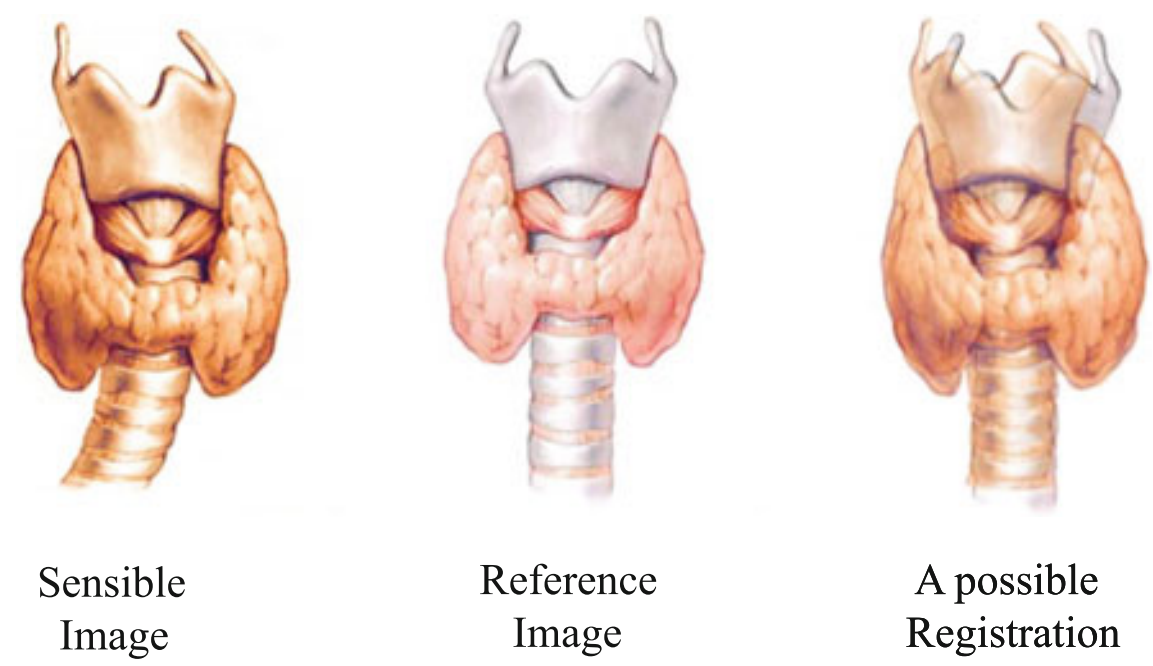

**Fig. 3** A registration result

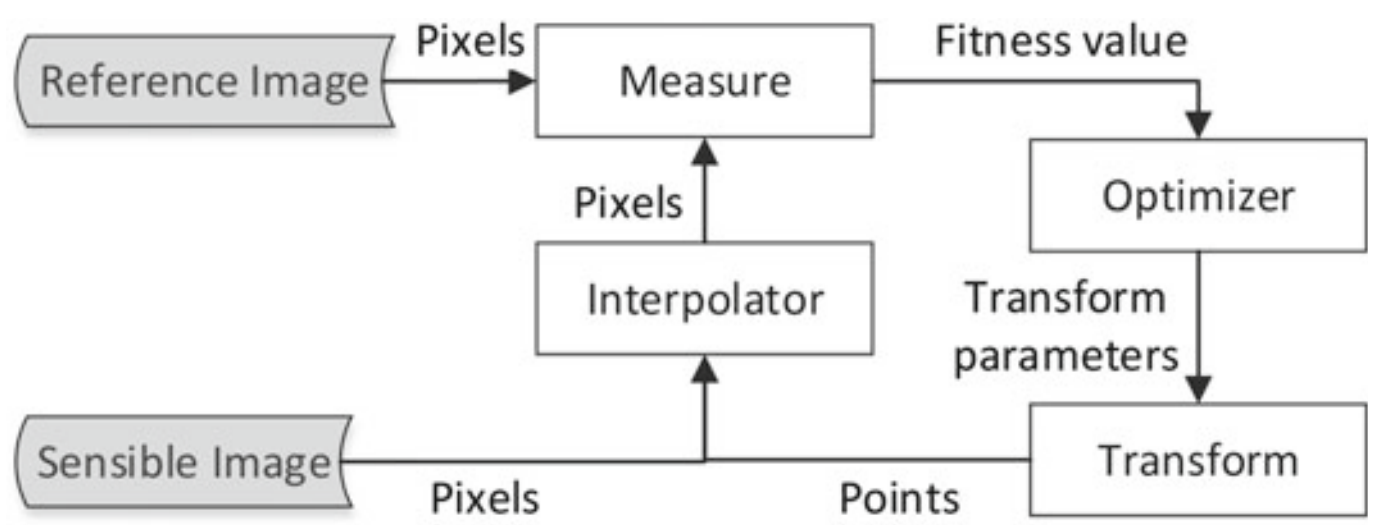

**Fig. 4** A registration process

Patients may undergo various magnetic resonance (MR), computed tomography (CT), single proton emission computed tomography (SPECT), and other imaging techniques for general or anatomical reference of a single organ. Therefore, physicians' analyses are benefited from registrations of images from any combination of modalities (including just a single modality). For instance, in radiotherapy treatments, both CT and MR can be used. CT is needed to compute the radiation dose accurately, while MR is usually better for delineating tumor tissues [11].

According to Maintz et al. [11], registrations are divided in three main categories: (1) the intrinsic, (2) the extrinsic and (3) nonimage based. In intrinsic registrations, the corresponding methods are based only on image-related content. The intrinsic registration can be based on the alignment of segmented binary structures or on a limited set of identified salient points (landmarks or keypoints).

In extrinsic registrations, artificial objects are attached to the patient, which are designed to be well visible and accurately detectable in all of the pertinent modalities. Therefore, the registration is comparatively fast and easy, which can virtually always be automated. Furthermore, since its parameters can be often computed explicitly, this kind of registration usually does not demand complex optimization algorithms.

Alternatively, registrations can also be nonimage based. If the imaging coordinate systems of the two scanners involved are calibrated to each other, then registrations can be performed automatically. In this case, scanners are usually required to be brought into the same physical location. Furthermore, it is also required to assume that the patient performs no movements between acquisitions.

Rigid transformations stand for a group of transformations where the distances between every pair of points are preserved [12, 13]. Therefore, rigid transformations comprise operations like translations and reflections, rotations, or combinations of these operations (some authors do not consider reflection as a rigid transformation). Affine transformations regard functions between affine spaces. These transformations preserve straight lines, points, as well as planes [14]. Operations such as scaling, translation, homogeneous and inhomogeneous dilations, reflections, shear mapping rotations and aggregations of these operations are defined as affine. The different types of transformations are depicted in Fig. 5. The overall range of transformations can either be (1) local, i.e., applied with regards to a certain region or (2) global, i.e., applied to the whole image.

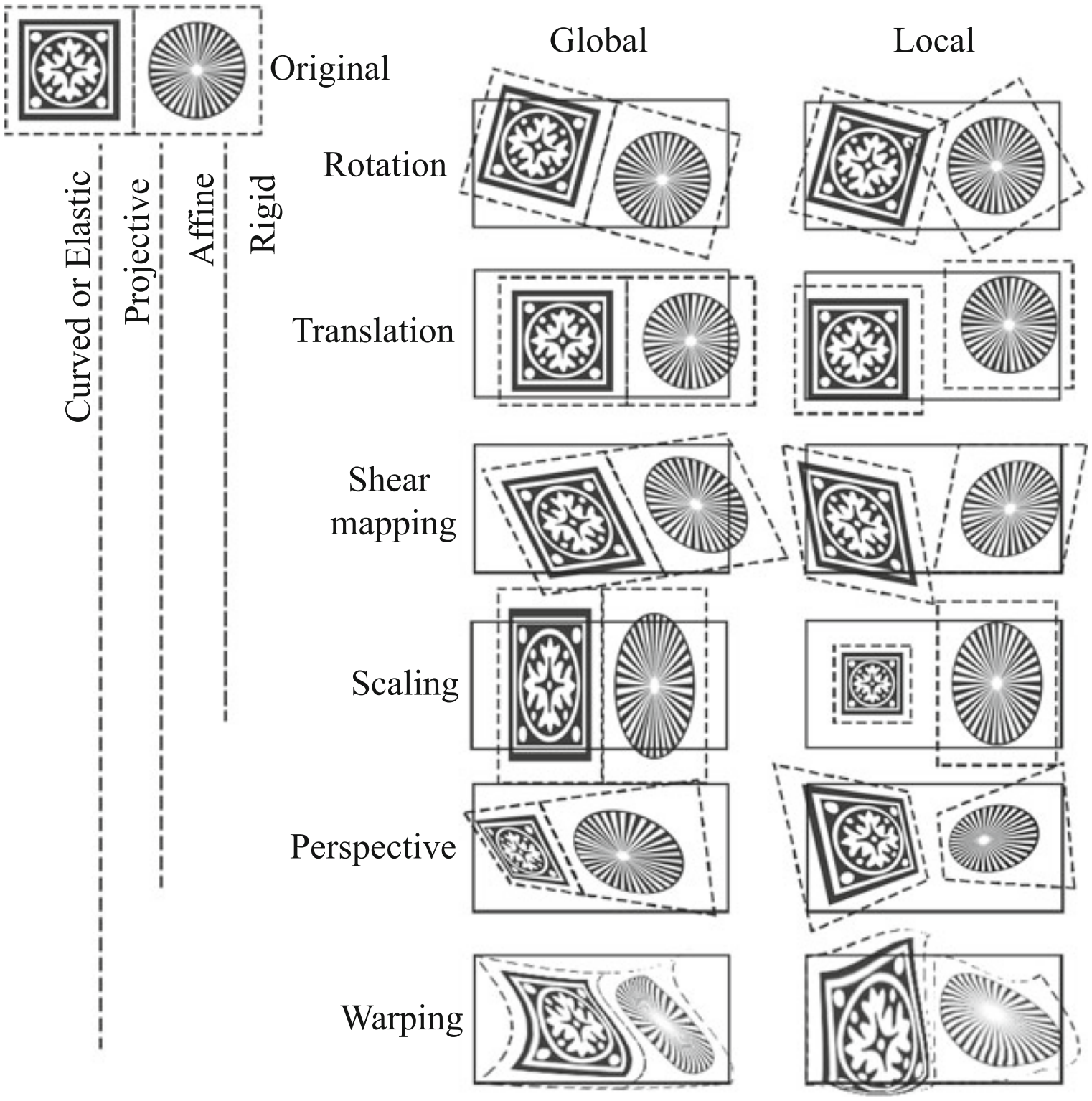

**Fig. 5** Possible types of transformations performed in registrations

In this work, we just consider rigid transformations, which in $\mathbb{R}^2$ are usually expressed as a linear matrix multiplication with homogeneous coordinates as shown in Eq. 1.

$$\begin{bmatrix} x' \\ y' \\ l \end{bmatrix} = \begin{bmatrix} \cos\theta & sen\,\theta & t_x \\ sen\,\theta & \cos\theta & t_y \\ 0 & 0 & 1 \end{bmatrix} \begin{bmatrix} x \\ y \\ l \end{bmatrix}, \tag{1}$$

where $(x', y')$ represents the transformed coordinates and $(x, y)$ the coordinates before the transformation. $\theta$ represents the angle of rotation and $t_x$ as well as $t_y$ the translation on the respective axes $x$ and $y$.

Common points of the reference and sensible images can be placed manually or autonomously to compose a landmark or keypoint registration approach. If this is the case, there must be a good precision in order to select the correct pixel in each image. However, it is difficult to manually place points when the number of required points is huge. For this case, some techniques can be used to autonomously detect points in the images such as the Scale-Invariant Feature Transform (SIFT) [15], Affine Scale-Invariant Feature Transform (ASIFT) [16], Speed Up Robust Features (SURF) [17], Harris Corner Detector (HARRIS) [18], etc. However, the disadvantage of these methods is that they often provide point correspondences that are not real. Besides, the amount of points can be huge, which may require some heuristic to discard and consequently reduce the number of points.

### *2.3 Thyroid-Related Review*

The thyroid is a neuroendocrine gland located at the human neck, next to the thyroid cartilage and over the trachea. The function of the gland is to regulate the metabolism of the body, produce proteins and to regulate the body sensibility with regards to other hormones. Some diseases are associated to the malfunction of the thyroid gland, which are usually related to iodine deficiency, such as goiter, hypothyroidism, thyroiditis, and random nodules that may be cancerous.

Thyroid nodules are a common clinical problem, and thyroid cancer is becoming more prevalent. According to the data provided by the American Thyroid Association in 2015 [19], papillary cancer can be the third most common type of cancer in women in 2019. In 2030, it can be the second most common in women and the third in men. Among the main types of cancer there are (1) the well-differentiated ones (papillary, follicular and the ones caused by Hurthle cells), (2) the medullar, and (3) the anaplastic.

In [20], the author observed that thyroid glands produce a very characteristic thermal pattern, which is of easy recognition. However, under hypoactivity, this pattern is not visible. Besides, this work reinforces that it is possible to discriminate benign nodules from malignant ones, according to the temperature pattern. It has also been observed that cancerous cells raise the temperature on neighboring regions. It also states that the more superficial the lesion, the higher is the temperature on the skin.

Clark et al. developed a prospective study using echography, ultrasound, and thermography on the preoperative evaluation of thyroid nodules [21]. Their objective was improving the accuracy on the differentiation between benign and malignant thyroid nodules, and determining the reliability of echography and thermography on distinguishing between solid and cystic thyroid nodules. The patients were in the supine position with the neck hyperextended. Sixty-one patients were analysed by clinical examination. By the combined use of echography and thermography, the authors were able to distinguish accurately between cystic and solid thyroid nodules.

In [22], D'Arbo et al. studied the usefulness of thermography in the selection of thyroid nodules for surgery. 124 thyroid nodules from 110 patients with ages in the range of 2–77 years old were studied. A hundred of these thyroid nodules were mapped as cold and the rest as hot. Sixteen cold nodules and two hot nodules were diagnosed as malignant. Each test took 15 minutes. Three different techniques were evaluated: heat index (related to a mean temperature of the region where the nodule was palpated), curve or thermal profile (established in respect to a middle point of the region) and, differences of the isotherms (related with the points of the higher temperature in the region compared to a near healthy area).

A design of a prototype device is proposed by Helmy and collaborators in [23]. It is an economical non-invasive system that detects and displays the relative skin temperature variations present in human patients suffering from thyroid disorders. In [24], the authors developed and described a model for thyroid glands, and simulated them using finite element analysis. This model was developed in order to determine the necessary resolutions for thermal sensors that are used to obtain the thermal images of patients' necks. In 2008, an improvement of this prototype was proposed [25]. The authors also present a finite-element analysis of a hot thyroid nodule. This analysis was used to investigate the temperature distribution.

In 2015, a successful computer model utilizing ANSIS software was proposed [26]. This simulation model incorporates three heat transfer coefficients: conduction, convection, and radiation. While the conduction component was a major contributor to the simulation model, the other two coefficients have improved the accuracy and precision of the model. This study also compares simulation data with the applied model generated from IR probe sensors. These data were analysed and processed to produce a thermal image of the thyroid gland. The acquired data were then compared with an Iodine uptake scan of the same patients.

In [27, 28], Gavriloaia et al. explain the details of a system used for acquiring infrared images of necks for thyroid nodules analysis. In these works, the Penne's equation and its applications were analysed. In [27], a few infrared images are explored with the aim of finding infrared signatures that can be descriptors of the thyroid tumors. The authors found that the contour of cancerous nodules is irregular and asymmetric. Using the ABCDE investigation method (based on features as Asymmetry, Border, Color, Diameter, and Evolution of the contour), 89.3% of the investigated patients with thyroid cancer were correctly recognized.

In [29], the authors apply fractal theory to quantify the irregularity in size and shape of thermal signatures of tumors. The self-similarity and lacunarity features were employed. In [30], an improved method for IR image filtering is proposed. This study is aimed at developing a numerical scheme that significantly reduces the computing time required for thermal image denoizing with edge preservation. This filter allows physicians to assess, faster than using other anisotropic diffusion filters, the contour shape, and to locate the outbreaks in ROIs. Another filtering is used in [31] to improve the quality of the information in thermographic images. In this case, the empirical mode decomposition was used, and it was proved that one or at most two intrinsic mode functions (IMFs) are enough for minimizing the noise effect.

The authors in [31] use thermography with other imaging modalities to diagnose thyroid problems. This work also evidences that female individuals are more susceptible to thyroid diseases than male and that hyperthyroidism is more common than hypothyroidism. The works [32, 33] employ an Otsu thresholding technique to separate the thyroid ROI from the rest of the images. In this work, we propose a novel registration and methodology to extract the ROI.

## 3 Proposed Approach

The images evaluated in this work (a single patient is shown in Fig. 6) were obtained using the FLIR ThermaCam S45 under the approval of the Ethical committee of Universidade Federal Fluminense, Brazil. The thermograms were acquired at the university hospital. All images were taken from patients and volunteers. These images were corrected for relative humidity and temperature of the room and were acquired after requesting patients to wait for 10 min in order to stabilize their metabolism. Four patients are analysed in this work (two healthy and two with pathological diagnosis or an abnormality in the thyroids). The blue square on the patient chin that can be seen in Fig. 6 is a physical marker that was placed at this region to help with the registration and recognition processes.

1. Protocol

With regards to the acquisition, we have used the dynamic acquisition protocol. However, it is slightly in regards to previous work due to the fact that it has a stopping condition for the thermal stress. Initially, the patient must be sitting down in order to minimize the possible displacements that he or she may perform. When the mean temperature of the skin decreases to 29 °C, the thermal stress is ceased and the sequential acquisition is started. One image is captured at every 15 s over 5 min, which produces the sequence of images shown in Fig. 6. The distance from the camera to the patient is 0.5 m, as shown in Fig. 1. The relative humidity of air and room temperature is recorded and inserted as parameters in the camera.

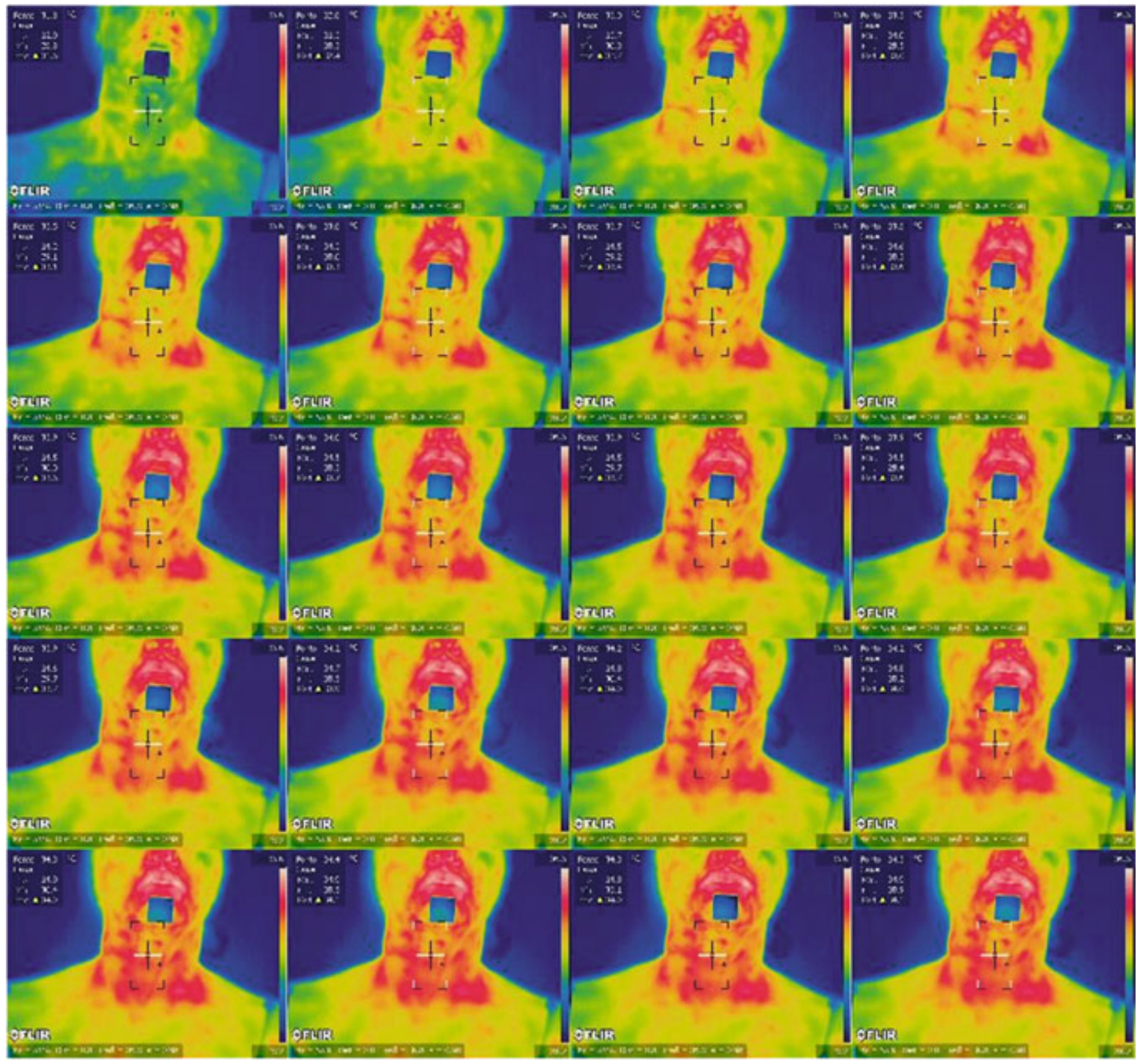

**Fig. 6** Sequence of 20 thermal images from an examination

2. Recommendations to Patients

At least 2 h before the examination the patient should avoid: alcohol, caffeine, physical exercises, nicotine and should not apply any cream, oil or deodorant.

3. Room Conditions

Room temperature is maintained between 22 and 25 °C, no windows, no openings, no air flow directed to the patient, and only fluorescent bulbs.

4. Preparation of the patient.

Inside the examination room, the patient is asked to remove earrings, necklaces or any other accessory that can be seen in the thermal image, central temperature is checked by a thermometer and hair stuck with a burrow. The patient must be in the room 20 min before starting the examination and should be sitting with his head tilted slightly back and looking up during acquisition.

## 3.1 Preprocessing—Movements During Acquisition

An important step in the processing of thermograms in the dynamic acquisition is registration. Figure 7 shows super-positions of two images of a single patient that moved during the acquisition (the first and the last one acquired in a sequence of 20 images). As it can be seen, the movements performed by the patients misalign the image content. By analyzing these movements, we can transform one of the images to resemble the other as much as possible. It can be done automatically or manually, and that usually involve transformations and similarity measures to compare the images, as previously addressed. As a remark, the green border of Fig. 7 precedes the red border.

Figure 8 compares the subsequent superposition of acquired images pairwise. It can be noted that the movements are not as severe as in Fig. 7. However, even slightly movements can impact on further processings. The extent of the movement does also vary depending on the patient.

The majority of the movements performed by patients can be described as follows:

1. Full-body Modification

Includes lateral movements to the left or right (Fig. 9a), movements to the top or bottom (Fig. 9b), or combinations of both movements. These tilts can be corrected with a rigid transformation: translation and rotation.

2. Local Modifications

These modifications include movements related to perspective and distortion, such as tilts of the head to the back or front. Figure 10 shows these tilts that disrupt the border of the head without altering the border of the shoulders so much. In these images, the chin of the patient was in a different position regarding the two acquisitions (red and green). These movements and distortions can be corrected with elastic (e.g., perspective, warping, etc.) transformations, which are more complex and more expensive (computational power-wise) than the rigid ones.

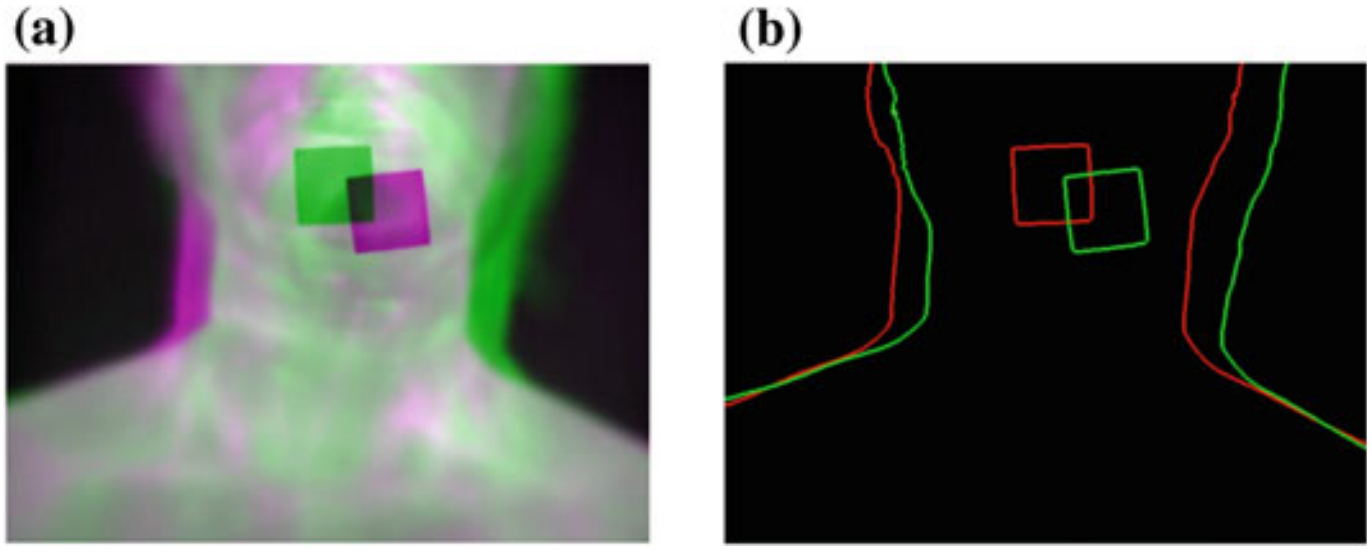

**Fig. 7** Movements of a patient during the image acquisition

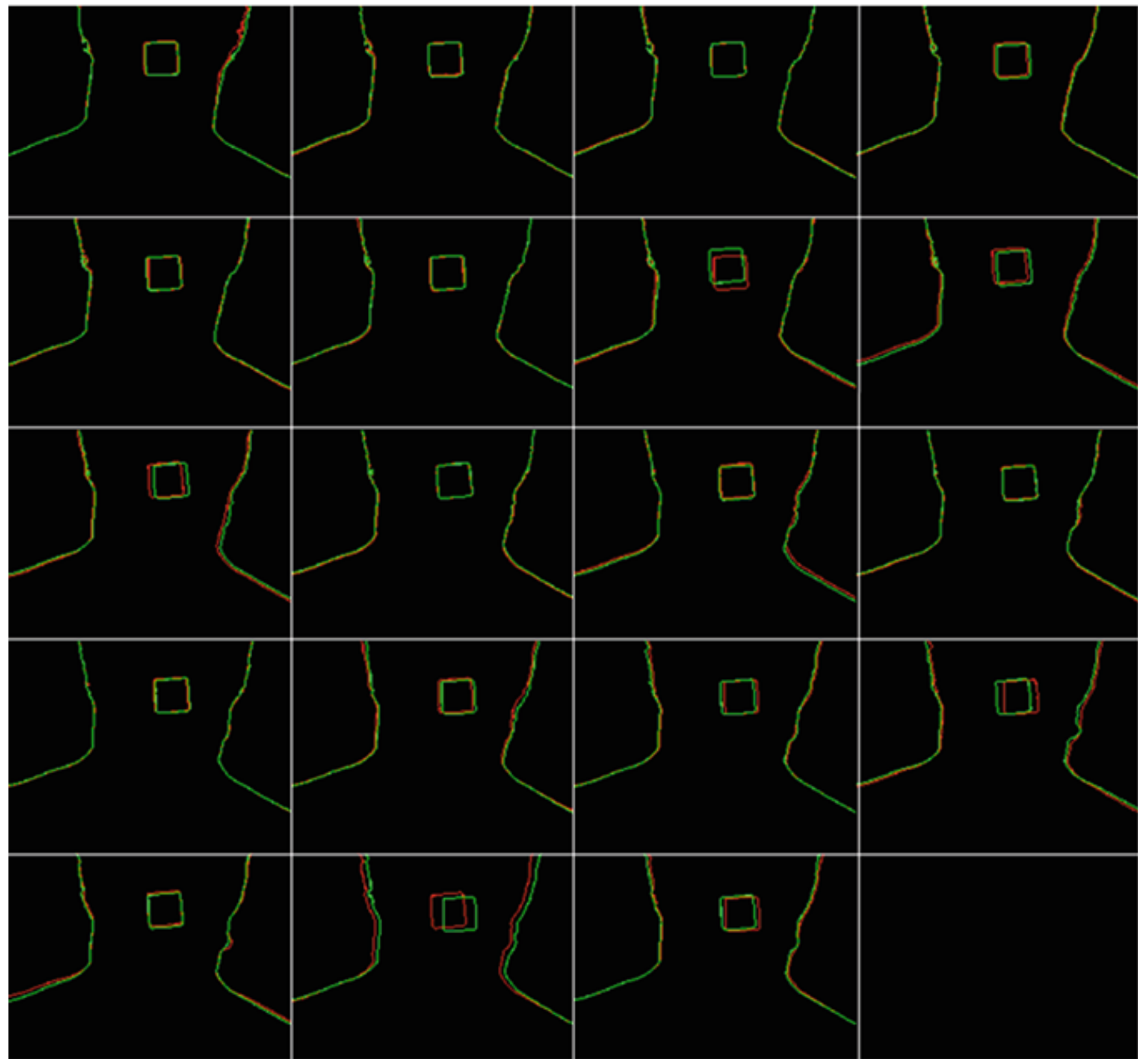

**Fig. 8** Subsequent movements on a time series acquisition

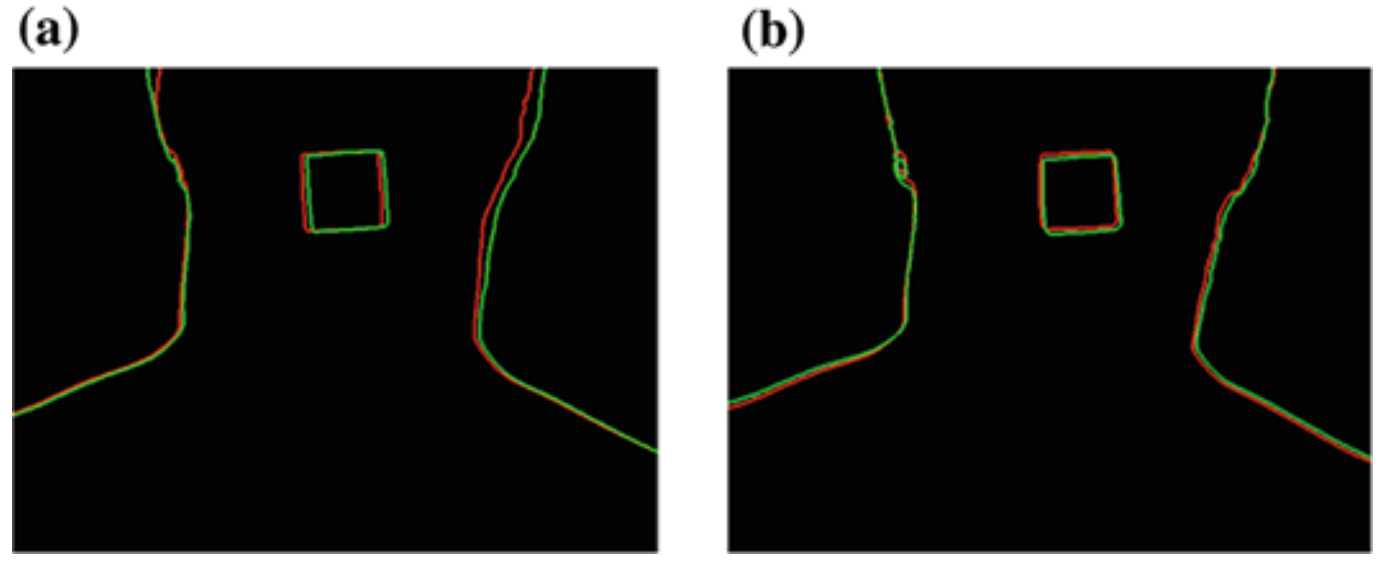

**Fig. 9** Slightly movements during the image acquisition

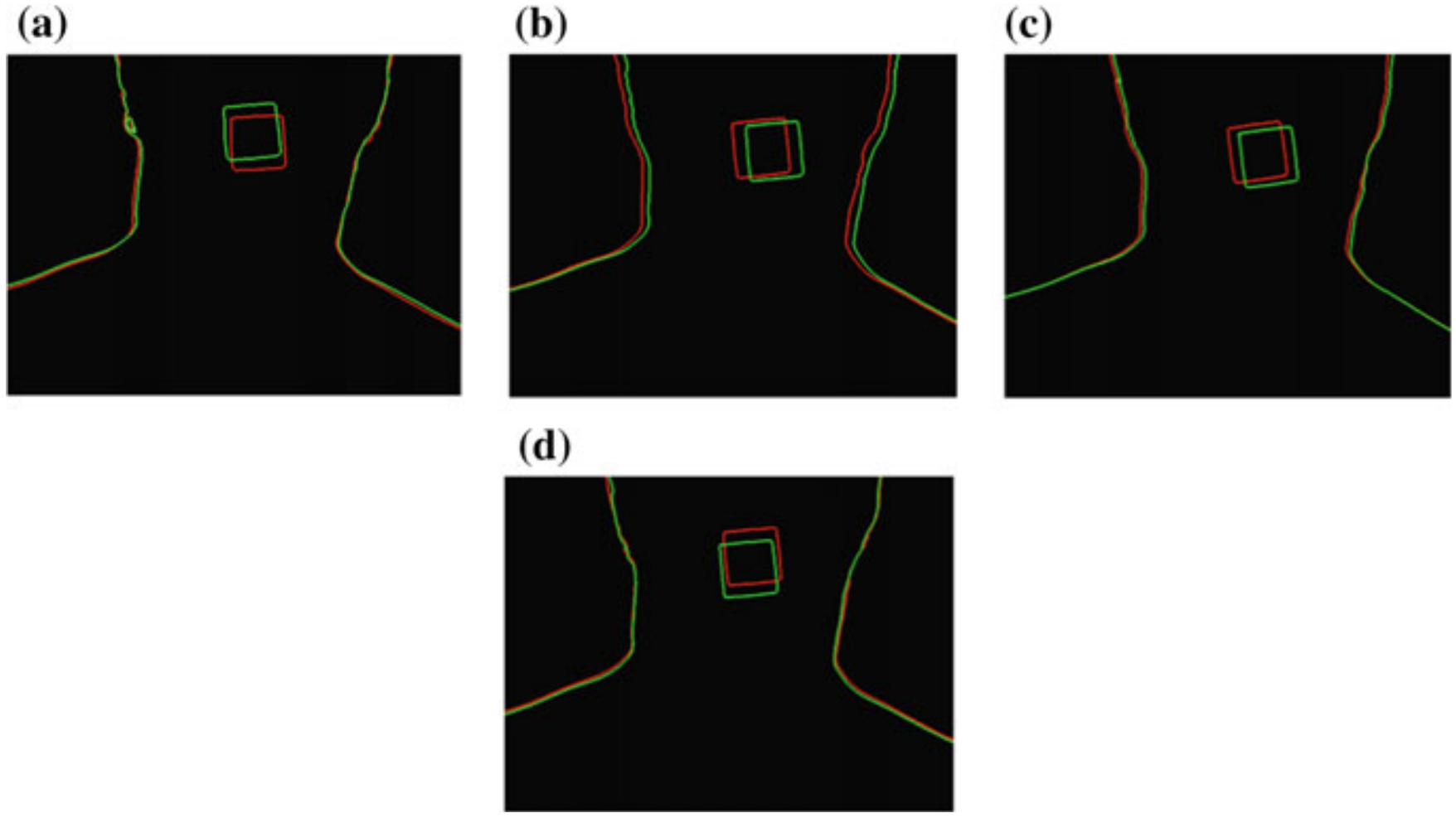

**Fig. 10** Complex movements during the image acquisition

## *3.2 Proposed Registration*

As previously discussed, registrations can be oriented by user intervention, where the user selects common points from both images. The advantage of such semi-manual registrations is that they are usually very accurate. However, at the same time they are time consuming and due to that fact it is not practical in several occasions. If some landmarks or keypoints are chosen on the subsequent images such as in Fig. 11, we can use this information to align the images by simply matching the points. Then, we would have something like Fig. 12.

It can be seen in Fig. 12 that the sequence of images is more aligned than in Fig. 8, since we have performed a semi-manual registration. It can also be noted that errors are present still. That is partially because we employed simple operations like translation and rotations only, and also due to the fact that the whole image was registered at once.

### 3.2.1 Autonomous Registration

Besides this semi-manual registration, we propose a possible autonomous registration for the thyroid thermographic images. At first, we apply a filter based on the Sobel operator, which was presented in a previous work [34] and is available at [35]. The filter is a modification of the classical Sobel filter, it is simple and considers kernels of variable sizes.

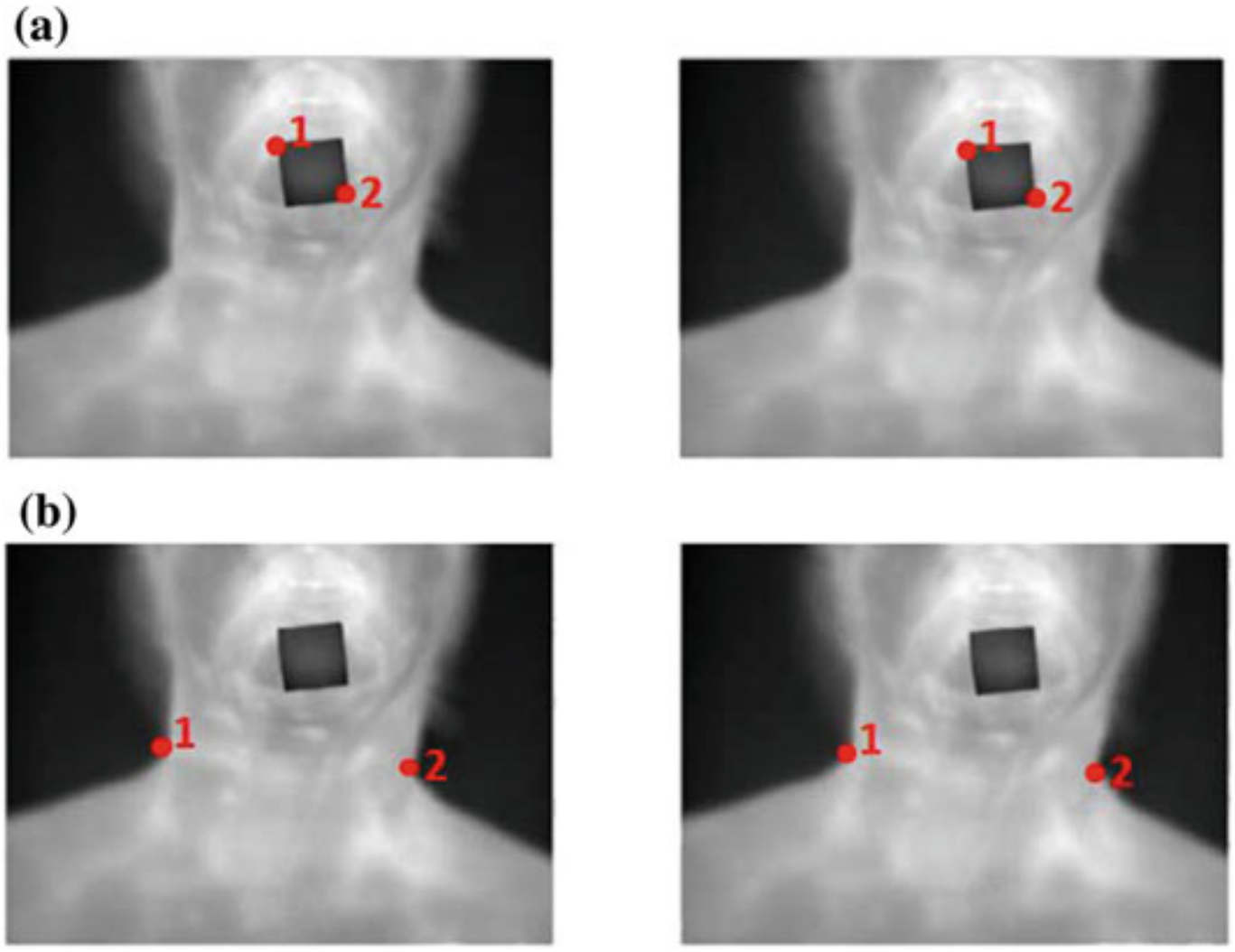

**Fig. 11** An example of landmark or keypoint selection

The Shadow and Light Sobel filters are defined by Eqs. 2 and 4, respectively, where $P$ represents the image and $P_{i,j}$ a value of a pixel at line $i$ and column $j$, $t$ represents a threshold for the edge to be painted, and $d$ a distance parameter (in classical Sobel operators $d$ would be close to 1), and $\wedge$ represents the boolean *and* operation.

$$\mathrm{SS}(P, d, t, i, j) = \begin{cases} 1, & \text{if } (s_1 \wedge s_2) \\ 0, & \text{otherwise} \end{cases} \tag{2}$$

where,

$$\begin{aligned} s_1 &= P_{i-d,j} - P_{i,j} > t \wedge P_{i+d,j} - P_{i,j} > t, \\ s_2 &= P_{i,j-d} - P_{i,j} > t \wedge P_{i,j+d} - P_{i,j} > t \end{aligned} \tag{3}$$

and

$$\mathrm{LS}(P, d, t, i, j) = \begin{cases} 1, & \text{if } (l_1 \wedge l_2) \\ 0, & \text{otherwise} \end{cases} \tag{4}$$

where,

$$\begin{aligned} l_1 &= P_{i,j} - P_{i-d,j} > t \wedge P_{i,j} - P_{i+d,j} > t, \\ l_2 &= P_{i,j} - P_{i,j-d} > t \wedge P_{i,j} - P_{i,j+d} > t \end{aligned} \tag{5}$$

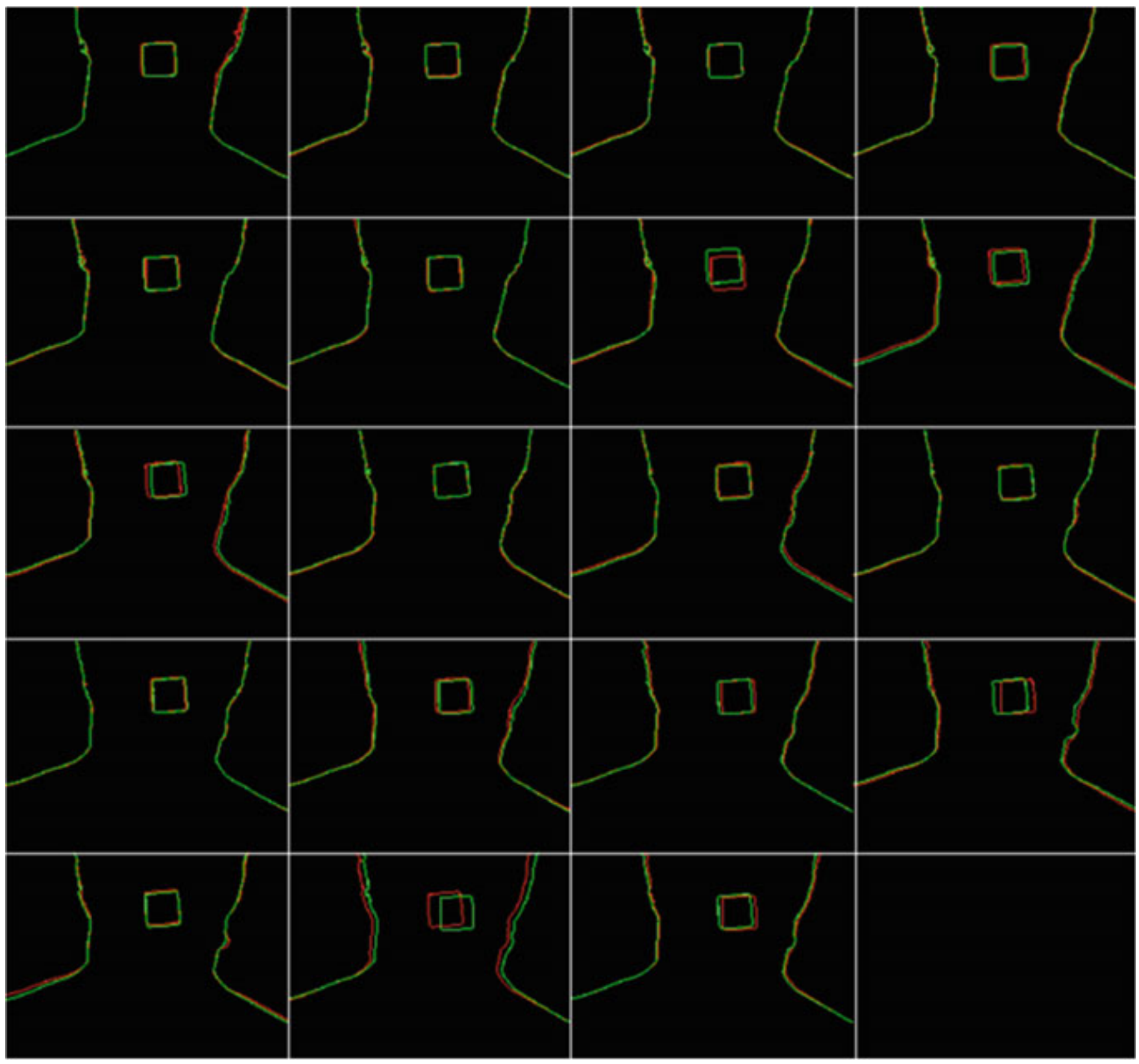

**Fig. 12** Registered subsequent images

The main difference between the variables $s_{1..2}$ and $l_{1..2}$ is the orientation of the computation, which captures either light or shadow valleys in each occasion. The parameters used in the proposed process were $t = -40$, and we have performed it only for the shadow and on the $x$-axis since it provided better results. That is, the LS function was disregarded as well as $s_1$. Thus, the function in Eq. 6 was applied to the images.

$$\mathrm{SS}(P,d,t,i,j) = \begin{cases} 1, & \text{if } (P_{i-d,j} - P_{i,j} > t \wedge P_{i+d,j} - P_{i,j} > t) \\ 0, & \text{otherwise} \end{cases} \tag{6}$$

Figure 13 shows the results of this filter applied to Fig. 13a. It can be noted that the lateral parts of the neck are clearly segmented and all the remaining stuff in the image is removed. From that image (Fig. 13b), it is very easy to cut off a rectangular shape of the area containing these lines, which would be our ROI.

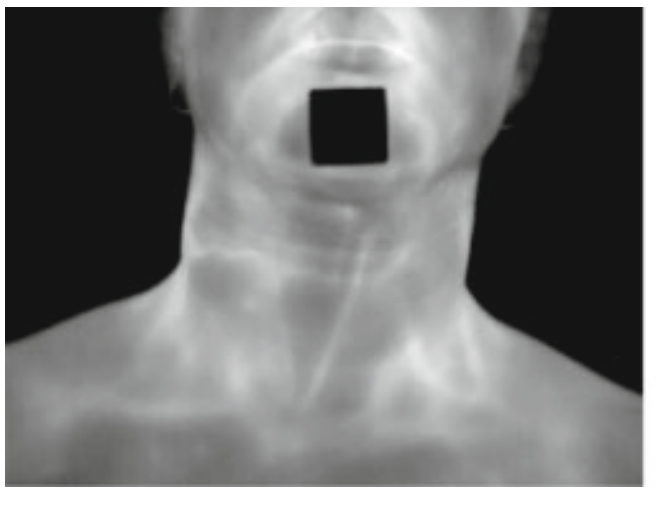

(a) Input image of a random patient.

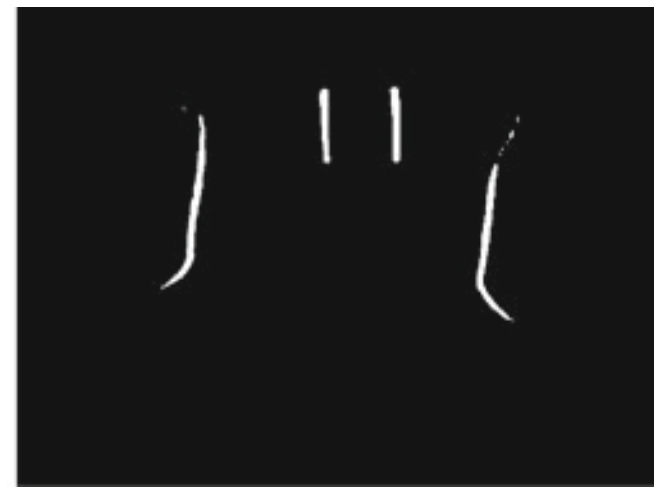

(b) Image $a$ after applying the filter.

**Fig. 13** Our Sobel-like filter

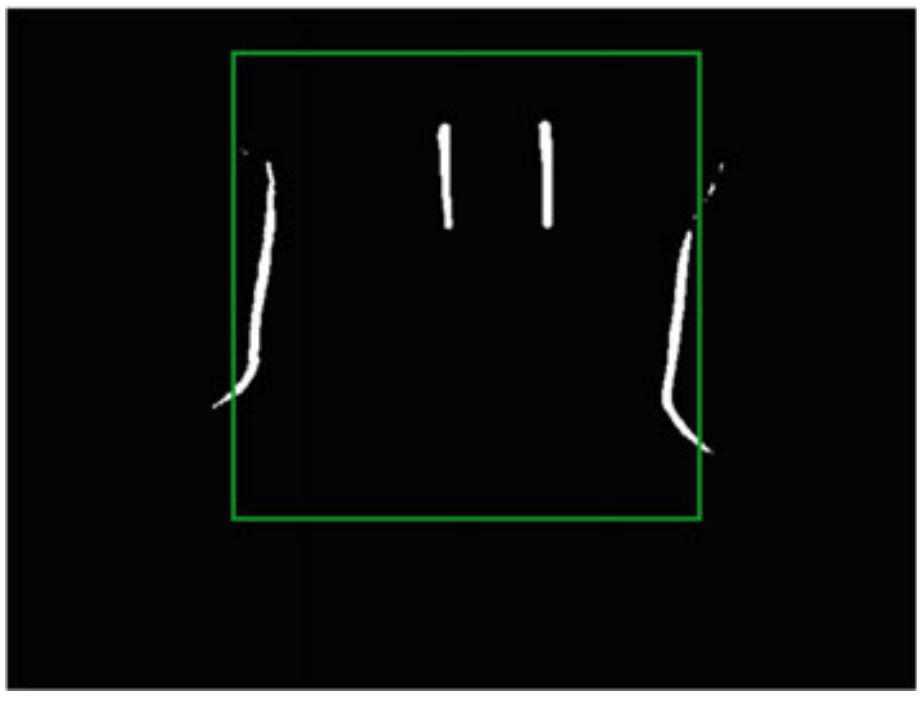

**Fig. 14** Autonomously detecting the ROI

Given a rectangle of fixed width and height (we have used 330 for width and 310 for height), a few candidate positions can be evaluated, and the one that contains the highest amount of white pixels would be the chosen ROI. Figure 14 shows an outcome of this described processing. The width or height of the rectangle can be reduced in order to produce different results.

It is clear by Fig. 14 that the two lines on the center of the image (which correspond to the physical marker) influence the position of the green rectangle. Therefore, the rectangle is forced upwards due to this residual "noise."

One would argue that we can erase these two central residuals by using the same heuristic described to find the ROI. And in fact we can. If we reduce the size of the rectangle and apply the same idea, we would find the central residuals. After that, we could just erase them completely from the image. Figure 15 shows the same idea when the size of the rectangle is reduced to $110 \times 110$.

After locating the residual noise we remove them. If the same heuristic is performed again, starting from bottom to top with regards to the $y$-axis, then we have the result shown in Fig. 16.

After extracting the ROI, we can use the different extracted ROIs to perform the registration. There are essentially two paths to follow. The ROI can be extracted before the registration or after the registration. In both ways, the described methodology using the Sobel derived filter would not change and would still be valid. However, there are some strengths and weaknesses on both of the approaches, which are discussed below.

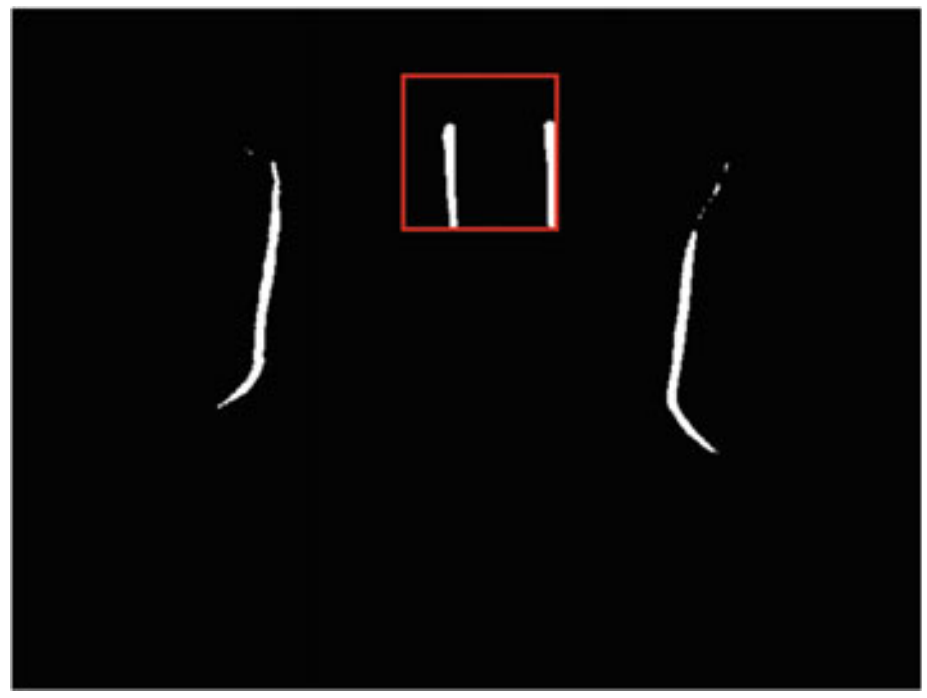

**Fig. 15** Removing the residuals

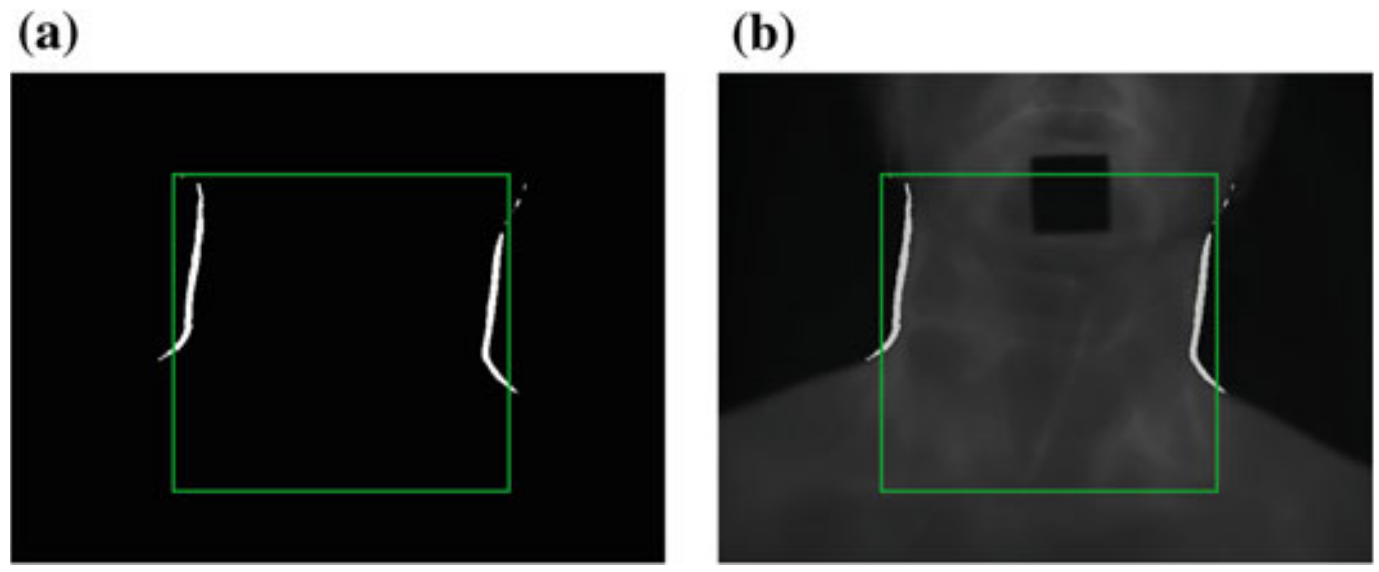

**Fig. 16** Final autonomous recognition of the ROI

1. ROI before Registration

In this case, just the neck region is considered for registration. The positive side of this is that the registration would probably be more accurate, since we would disregard the information of the shoulders and other structures. For instance, we have observed that some patients perform movements with their heads while their shoulders stand still. If the head is moved, so is the neck. In that way, if the shoulders are regarded in the registration, then they would influence the final registration result. And since the shoulders are already aligned, the registration would not change the image significantly.

The disadvantage of this approach is that the comparison of the images would have to be done regarding the intensity values of the ROIs, which is slightly slower to compute than to consider a few keypoints of both images. However, it is still very fast, lasting only a few milliseconds for a pairwise comparison. As a remark, the intensity values can also vary considerably from one image to another, so that a proper similarity measure should be employed. A simple similarity measure such as the mean difference would probably achieve a bad performance. The Chamfer distance [34] would probably be more accurate.

2. ROI after Registration

In this case, it is easier to find common landmarks or keypoints between the images, since the shoulders and other structures are still present in the image. However, it would introduce the errors of the shoulders and other structures into the registration, which should be for the neck only in the first place.

Furthermore, no information is disregarded before the registration, so that it can be used for any further processing if required, even to help with the diagnosis, as opposed to the previous paradigm.

## 3.3 Feature Analysis and Classification

We have also noticed that it is possible to locate the nodule of sick patients with a simple threshold operation (which was confirmed by a specialist on the available images). However, we are not sure if this is true for a wider group of diseased patients although it seems so. A threshold operation takes the image as input and changes the pixel or temperature value to white if it is higher or equal than a threshold value or changes it to black if it is lower. The threshold value was set to 209, so, values higher or equal than 209 become white and values lower than that become black.

Given the thresholded images in Fig. 17, it is already possible to extract some valuable information from them. Three features directly based on the area segmented by the threshold operation along with a symmetrical feature were analysed. The features based on the segmented area are computed on the basis of all the pixels within the segmented hot region.

The features based just on the thresholded area are (1) the mean intensity, (2) the standard deviation of these intensities and (3) the maximal intensity or temperature. The fourth feature quantifies the level of symmetry of each ROI in each image, with respect to the vertical axis, and is based on the idea that if two pixels belong to the same symmetric zone then their intensities should be the similar. Thus, the absolute difference between these values should be minimal.

We analyse the intensity values of all the pixels of the ROI in the following manner: let $m$ and $n$ be the height and width of the ROI, $n/2$ be half of the ROI rectangle width (due to the vertical symmetry axis) then, for each pixel $(i,j)$ of the left part of the ROI, we look for the most similar intensity value in the neighborhood $(3 \times 3)$ of the pixel $(i, n-j)$, which is located at the right part of the ROI. Similarly, for each pixel on the right side, a similar intensity is searched for in the

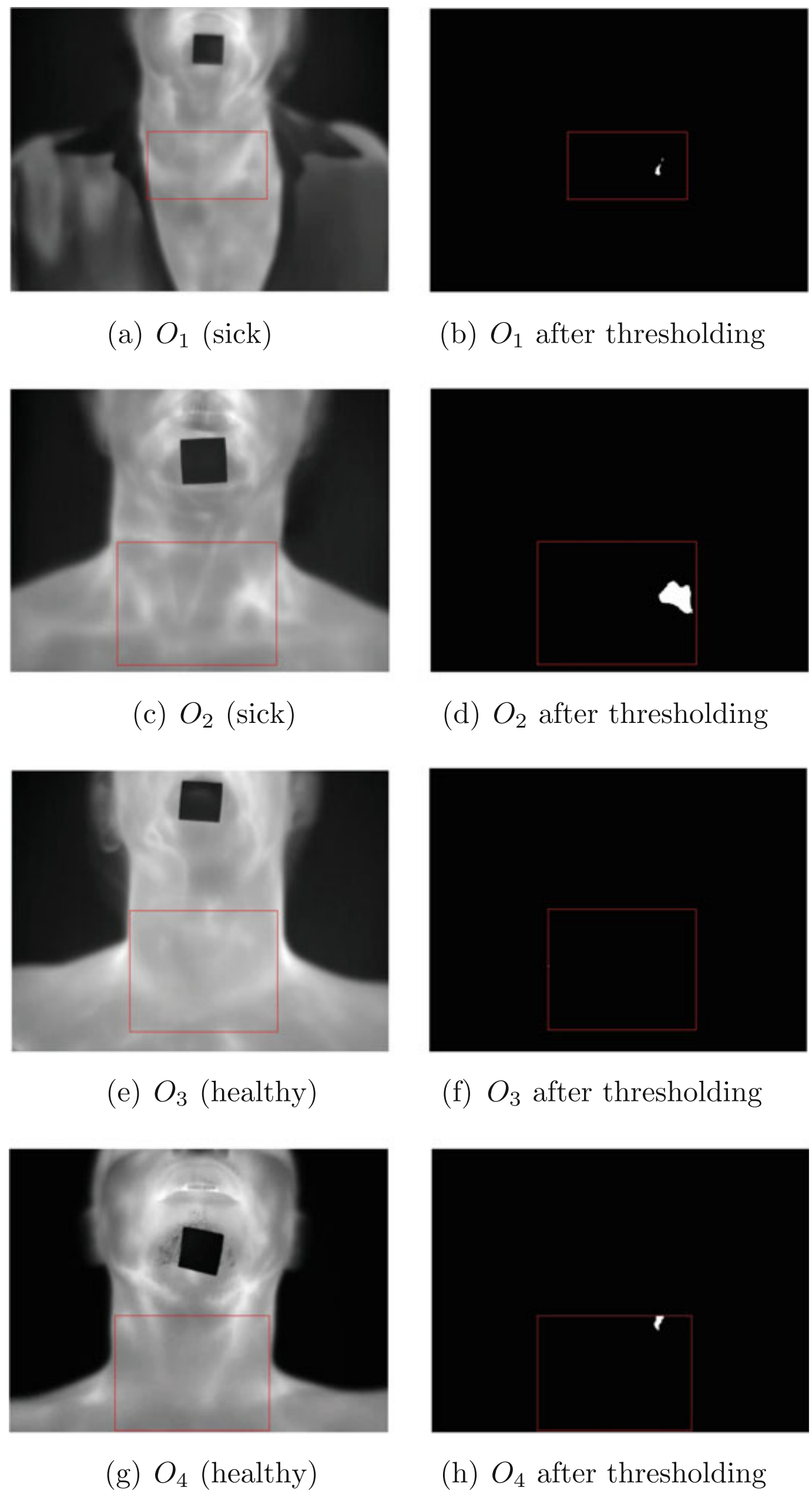

(a) $O_1$ (sick) (b) $O_1$ after thresholding

(c) $O_2$ (sick) (d) $O_2$ after thresholding

(e) $O_3$ (healthy) (f) $O_3$ after thresholding

(g) $O_4$ (healthy) (h) $O_4$ after thresholding

**Fig. 17** Temperature analysis

left part in the same fashion. The asymmetry level is given by Eq. 7, which stands for the average sum of the absolute values of the intensity difference of each pixel and its most similar one. The difference between each pair of pixels is normalized regarding the highest intensity value (in this case, 255).

$$\mathrm{Asy}(I) = \frac{1}{mn}\left(\sum_{i=1}^{m}\sum_{j=1}^{\frac{n}{2}} \frac{\min_{p,q=-1:1} |I(i,j) - I(i+p, n-j+q)|}{255} + \sum_{i=1}^{m}\sum_{j=\frac{n}{2}+1}^{n} \frac{\min_{p,q=-1:1} |I(i,j) - I(i+p, n-(n-j)+q)|}{255}\right) \quad (7)$$

We have employed the $k$-Nearest Neighbor ($k$-NN) algorithm to perform a classification using the extracted features, where the Euclidean distance was regarded and $k$ was equal to 1. As previously mentioned, two patients contain nodules ($O_1$ and $O_2$) while the other two do not ($O_3$ and $O_4$). Table 1 shows the Euclidean distances of patient $O_1$ in relation to all the others. It can be observed that patient $O_1$ is tightly related to patient $O_2$, which makes sense, because both of these patients are sick.

Each patient has four feature values. For each patient, we take their four feature values and compute their distance in regards to all the feature values of the remaining patients. For instance, the distance from the standard deviation of $O_1$ is calculated in relation to the standard deviation of $O_2$, $O_3$ and $O_4$. That calculation is performed for each feature (normalized mean intensity, standard deviation, normalized maximum temperature and asymmetry). For each pair of patients we have a total of four distance values, which represent the four features.

Thus, the lesser the distance value between patients $O_k$ and $O_l$, the closer patient $O_k$ is in relation to patient $O_l$ and vice-versa. Therefore, in Table 1, the minimal distances of the four features are all related to $O_2$, which indicates that $O_1$ is tightly related to $O_2$. In Table 2, three features indicate that $O_3$ is related to $O_4$, so that we assume that its class is the same class as of $O_4$, which is healthy. For the patient $O_2$, two of its features indicate similarity with $O_1$, while the remaining two were closer to $O_3$. Three features indicate that $O_4$ can be classified as healthy.

**Table 1** Euclidean distance of patient 1 ($O_1$) in relation to the remaining

| Feature | $O_2$ | $O_3$ | $O_4$ |
|---|---|---|---|
| Normalized mean intensity | **0.172** | 0.8216 | 0.414 |
| Standard deviation | **14.648** | 28.826 | 37.26 |
| Normalized maximum temperature | **0.157** | 0.833 | 0.315 |
| Asymmetry | **0.053** | 0.063 | 0.070 |

Bold values indicate the nearest value, or the smallest difference between the features of patient $O_1$ and the remaining

Table 2 Euclidean distance of patient 1 ($O_3$) in relation to the remaining

| Feature | $O_1$ | $O_2$ | $O_4$ |
|---|---|---|---|
| Normalized mean intensity | 0.821 | 0.848 | **0.811** |
| Standard *d*eviation | 28.826 | 20.963 | **15.977** |
| Normalized maximum temperature | **0.833** | 0.898 | 0.971 |
| Asymmetry | 0.063 | 0.014 | **0.006** |

Bold values indicate the nearest value, or the smallest difference between the features of patient $O_3$ and the remaining

# 4 Conclusion

This chapter reports a small scale preliminary study that evaluates the feasibility with regards to time, cost, and effect of the use of infrared imaging as a tool in the detection and analysis of thyroid nodules. We perform an overall analysis of thermographic images, focusing on thyroid thermographic acquisition, processing and classification, which is a new field of study. We discuss the data and different approaches with regards to a computer science background, in order to extract, make sense of the information captured by these images, and eventually predict or provide automatic diagnoses.

An autonomous ROI identification for the thyroid images is proposed, which is based on very simple fundamentals of computer vision. We have discussed different approaches with regards to the registration and ROI extraction of these images, highlighting the advantages and disadvantages of each one. More specifically, we discuss whether it is better to extract the ROI before the registration rather than after.

Furthermore, we have also performed an extraction and analysis of four features from the ROI of the patients using a classification algorithm (*k*-NN). In our analysis, we found some evidence that these features may be sufficient to predict whether a patient is sick or not.

Future work involve improving the described methodologies and analysing the reported evidences with regards to a large scale of data. Since there is still a lack of thermographic thyroid images with abnormalities in our repository, we were forced to work with the currently available images. Still, we expect that the evidence found in this work applies to the general case.

**Acknowledgments** The authors would like to thank to CAPES, CNPq (Project: N0. 201542/2015-2 PQ—CA: EM), FAPERJ and FAPEMA, Brazilian agencies, for partially funding this work. This research has also been partially supported by projects INCT-MACC and SiADE.